# Enhancing Knowledge Entity Relation Extraction in Abstracts and Full-text Content of Academic Articles Using Large Language Models

**Yingyi Zhang [a], Chengzhi Zhang [b, *]**

[a] Department of Archives and E-government, Soochow University, Suzhou, China

yyzhang9@suda.edu.cn

[b] Department of Information Management, Nanjing University of Science and Technology, Nanjing, China

zhangcz@njust.edu.cn

**Abstract:** This study investigates the effectiveness of large pre-trained language models (LLMs) for knowledge entity relation extraction (KERE) from academic papers. Existing studies have examined individual types of LLM-based strategies and evaluated model performance using metrics such as precision and recall. To provide a broader evaluation, this study investigates multiple strategies, including closed-source LLM-based CoT prompting, LoRA-adapted open-source LLMs, LLM-enhanced small pre-trained language models (SLMs), and SLM-enhanced closed-source LLMs. In addition, this study introduces two strategies, namely SLM-enhanced open-source LLMs and closed-source LLM-enhanced open-source LLMs. These strategies are evaluated from multiple perspectives, including metrics, their ability to mitigate errors observed in SLMs, and potential directions for model optimization. This study manually constructs two annotated datasets in the NLP domain, covering Abstract and Full-Text datasets. Experimental results show that the LoRA-adapted T5 model achieves the highest macro $F_1$ scores of 0.7408 and 0.7013 on the Abstract and Full-Text datasets, respectively. SLMs also benefit from LLM-generated entity descriptions under the LLM-enhanced SLM strategy. Error analysis further reveals that the effectiveness of LLM-based strategies in mitigating SLM errors varies across error types. Based on the identified limitations and error patterns of these strategies, several optimization directions are proposed. In addition, combining strategies with different error patterns further improves KERE performance.

## 1. Introduction

Academic papers contain various knowledge entities, such as research problems, methods, datasets, and evaluation metrics, together with the relations among them. Knowledge entities are the minimal independent units that summarize crucial concepts and essential elements in scientific literature (Zhang et al., 2025). Knowledge entity relation extraction (KERE) is a classification task and serves as a prerequisite for constructing triples between knowledge entities (Zhang et al., 2017). Accurate extraction of entity relations supports downstream applications, including knowledge graph construction (Luan et al., 2018), academic impact assessment (Luo et al., 2022), and knowledge evolution analysis (Zhang et al., 2022).

* Corresponding author: Chengzhi Zhang.

Large pre-trained language models (LLMs) have recently been applied to various academic information extraction tasks, including citation context extraction (Nguyen et al., 2026), sentence extraction (Zhang & Zhang, 2024), and knowledge entity recognition (Liu et al., 2025). Following the categorization adopted by Muennighoff et al. (2023), this paper refers to models with more than one billion parameters as LLMs and models with fewer than one billion parameters as small pre-trained language models (SLMs). For KERE, existing studies have explored several LLM-based approaches, including chain-of-thought (CoT) prompting (Ma et al., 2023), LoRA-adapted open-source LLMs (Zhan et al., 2025), and LLM-enhanced SLMs (Asada & Fukuda, 2024), particularly in domains such as biomedicine and materials science. Existing studies have also compared different LLM-based strategies, with most comparisons focusing on variations within a single methodological category, rather than systematically comparing different LLM-based strategies. For example, Peng et al. (2026) compared the performance of encoder-only and decoder-only LLMs, as well as full fine-tuning and LoRA. Similarly, Al Nazi et al. (2025) evaluate open and closed-source LLMs with differences prompting techniques, such as zero-shot, few-shot, and CoT prompting.

This study compares five representative LLM-based strategies for KERE. Specifically, this study evaluates three existing strategies, including closed-source LLM-based CoT prompting, LoRA-adapted open-source LLMs, and LLM-enhanced SLMs, and further investigate two additional strategies: SLM-enhanced LLMs and closed-source LLM-enhanced open-source LLMs. In the SLM-enhanced LLM strategy, an SLM serves as a relation-type filter by predicting the probability distribution over candidate relation types and selecting the top-N relation types with the highest probabilities to construct prompts that guide the LLM's reasoning process. This strategy has been applied to Chinese entity relation extraction (Tang et al., 2024). Building on the approach of Tang et al. (2024), who incorporated the top-N relation types into prompts for closed-source LLMs, this study further leverages the filtering results to guide LoRA-adapted open-source LLMs. In addition, the closed-source LLM-enhanced open-source LLM strategy employs closed-source LLMs for relation explanation generation and training sample quality evaluation. Closed-source LLMs are proprietary models that are not publicly available for adaptation, whereas open-source LLMs are freely accessible and can be adapted for downstream tasks (Al Nazi et al., 2025). In this strategy, closed-source LLMs are accessed via APIs, while open-source LLMs are adapted to domain-specific datasets using LoRA. Due to hardware limitations during LoRA adaptation, this study select open-source LLMs with relatively smaller parameter sizes. Therefore, this strategy aims to combine the domain-specific knowledge learned by LoRA-adapted open-source LLMs with the broad general knowledge encoded in larger closed-source LLMs.

Systematically evaluating different LLM-based strategies for KERE requires high-quality benchmark datasets. With the rapid advancement of LLMs, their powerful language understanding capabilities have driven transformative changes across various fields, and these advancements are largely enabled by progress in natural language processing (NLP) (Radford et al., 2018). In the NLP domain, extracting relations between entities from academic literature can support various downstream applications, including technology development revealing (Zhang et al., 2024a) and future skills identification (Gonzalez-Gomez et al., 2025). Existing manually annotated datasets mainly focus on general relation types, such as used-for, part-whole, compare, and hypernym-hyponym (Gǘbor et al., 2018; Luan et al., 2018), leaving several relation types with important downstream applications underexplored. For example, compared with the existing undirected "compare" relation, a directional "compare" relation between two method entities can provide detailed information for the identification of technological evolution paths (Xue & Shao, 2024). To address these limitations, this study constructs two manually annotated datasets, namely the Abstract Dataset and the Full-Text Dataset, in the NLP domain. During dataset construction, the relation schema was designed by combining existing relation types identified from previous studies with new relation types discovered through the analysis of NLP academic papers. As mentioned above, the first research question is posed:

***RQ1:*** How do different LLM-based strategies perform on knowledge entity relation extraction across the two datasets?

In comparative experiments evaluating the performance of different LLM-based methods, traditional metrics such as precision and recall are widely used (Asada & Fukuda, 2024; Peng et al. 2026) . Although these metrics reflect overall extraction performance, they are insufficient to reveal the specific strengths of models in the complex task of KERE. To overcome this limitation, this study conducts detailed manual annotation and error analysis of relation extraction results produced by SLMs such as SciBERT, and systematically analyzes how various LLM-based methods address these issues to uncover their differentiated strengths. Therefore, the second research question is posed:

***RQ2:*** To what extent can different LLM-based KERE strategies mitigate the errors observed in SLM-based KERE models?

Although the above analysis reveals the strengths of LLM-based strategies over SLMs, it does not clarify the remaining limitations of LLMs themselves. To further investigate these limitations, this study applies the automated error analysis framework proposed by Das et al. (2022) to analyze the distribution of error types across different LLM-based strategies and identify potential directions for improvement. As mentioned above, the third research question is posed:

***RQ3:*** What limitations do different LLM-based strategies exhibit for KERE, and how can these limitations be addressed through targeted optimization?

The main contributions of this study are summarized as follows. First, this study constructs two manually annotated datasets for KERE, including the Abstract dataset and the Full-Text dataset. By combining literature review and data analysis, this study develops a relation schema and introduces three additional relation types. Second, this study compares five representative LLM-based strategies for KERE and investigates two additional strategies, i.e., SLM-enhanced LLMs and closed-source LLM-enhanced open-source LLMs. The results reveal the performance differences of different LLM-based strategies. Third, this study conducts error analyses to examine the strengths and limitations of different LLM-based strategies. The identified error patterns indicate possible directions for improving LLM-based KERE models.

The data sample and source code of this paper are freely available at the GitHub website: https://github.com/YingyiZhang/Knowledge-Entity-Relation-Extraction-with-LLM.

## 2. Related work

The aim of this paper is to extract knowledge entity relationships from academic papers and to compare the performance of different LLM-based strategies on two manually annotated datasets. To achieve this objective, this section reviews the existing literature from two perspectives: KERE methods for academic papers and existing KERE datasets.

### 2.1 Knowledge entity relationship extraction methods for academic papers

Currently, the mainstream methods in this field are primarily based on pre-trained language models（PLMs）. According to the difference in model size, PLMs can be categorized into small pre-trained language models, such as BERT and SciBERT, and large pre-trained language models, such as GPT-4 and Llama-2. This section will systematically review the methods related to KERE from academic papers based on these two types of models.

#### (1) KERE with small pre-trained language models

In the methods using SLM, Zhong & Chen (2021) compared the performance of BERT and SciBERT in KERE, finding that SciBERT outperforms BERT. To improve the performance of entity relation extraction tasks, existing research has optimized model structures to address issues such as coreference resolution errors, relation direction prediction errors, and entity contextual semantic understanding errors. For example, to address coreference resolution errors, Luan et al. (2018) introduced a mention ranking module specifically for identifying noun phrases referenced by pronouns. To mitigate relation direction prediction errors, studies have modeled the dependencies between entities and their relations. Xu et al. (2024) used self-attention mechanisms to learn the semantic dependencies between entities, while Chen et al. (2022) used graph convolutional networks to learn these dependencies. To improve entity contextual semantic understanding, researchers have incorporated various features of entities and their contextual information to enhance their semantic representation. For instance, Aid et al. (2024) employed SciBERT to encode entity representations and integrate them with sentence representations, whereas Yang et al. (2025) employed convolutional neural networks to automatically filter features for entity relation extraction, such as entity type and part-of-speech.

In addition to extracting relations based on known entities, some studies proposed joint models that simultaneously perform entity recognition and entity relation extraction. For example, Eberts & Ulges (2020) first classify each span in the sentence and treat specific spans as knowledge entities, which are then passed to the relation extraction module for relation classification. Mullick et al. (2024) used Pointer Networks to identify entities in the sentence and then input these entities into the relation classification module. Xu et al. (2026) combined entity and relation extraction with a self-attention mechanism to generate rich feature representations through multi-task dynamic interaction.

#### (2) KERE with large pre-trained language models

In methods using LLMs, existing research has attempted to apply prompt learning, LLM adaptation, LLM-enhanced SLMs and SLM-enhanced closed-source LLMs to perform KERE in academic texts.

In the prompt learning approach, Dagdelen et al. (2024) used prompt learning to extract compositional and application relationships between entities in the field of material chemistry. Some studies have optimized prompt learning methods. For example, Efeoglu & Paschke (2024) proposed a retrieval-augmented generation method, which enhances prompt information by searching for similar samples from the annotated dataset. Laskar et al. (2025) designed two stage methods, first is to use LLM to extract relation, second is to design a prompt to use LLM to judge the relation extraction result.

In LLM adaptation methods, several research institutions have open-sourced their LLMs, such as Llama (Touvron et al., 2023), ChatGLM (GLM et al., 2024), Baichuan (Yang et al., 2023), and Deepseek (DeepSeek-AI et al., 2025) model. Han et al. (2024) pointed out that adaptation is necessary to equip LLMs with domain adaptation capabilities. Hu et al. (2022) proposed the LoRA (Low-Rank Adaptation of LLMs) framework, which trains only a portion of the model's parameters while achieving performance close to fine-tuning all parameters. Dagdelen et al. (2024) employed LoRA to adapt Llama-2 and other LLMs for extracting relationships between entities like metals and materials from academic papers in material chemistry. Similarly, Collins et al. (2026) evaluated the performance of LLaMA 3.1 with LoRA for relation extraction tasks in the biomedical domain. It is important to note that these studies fixed the prompt templates during adaptation, meaning the input prompt information could not be changed during the process. In contrast, some studies introduced soft prompt by embedding trainable vectors into the prompt, allowing for task adaptation through optimization. For example, some studies used soft prompts to adapt LLMs on clinical and biomedical texts, with results indicating that adaptation with soft prompts slightly outperformed methods using fixed prompts (Peng et al., 2024; Zhou et al., 2024).

In LLM-enhanced SLM strategies, Asada & Fukuda (2024) proposed two approaches to integrate LLMs with SLMs. The first approach used LLMs to generate explanations of the relations in the training dataset, while the second approach used LLMs to generate descriptions of entities, supplementing the original training data. Experimental results from biomedical abstracts showed that the second approach could enhance the performance of SLMs. In another study in the agricultural field, GPT-3 and other large language models were used for data augmentation, and the generated high-quality samples were added to the original training set to fine-tune SLMs (De et al., 2025). In the SLM-enhanced closed-source LLM strategies, Tang et al. (2024) used an SLM as a relation type filter by predicting the probability distribution over candidate relation types and selecting the top-N relation types with the highest probabilities to construct prompts that guide the LLM's reasoning process.

### 2.2 Dataset for knowledge entity relationship extraction task

The datasets annotated for KERE can be categorized into Abstract datasets and Full-Text datasets, which are introduced as follows.

#### (1) Abstract datasets for knowledge entity relationship extraction task

Abstract datasets for KERE are constructed from the abstracts of academic papers. For example, Tateisi et al. (2016) constructed a dataset comprising 400 abstracts in the computer science domain, with annotations for relations such as used-for, part-whole, compare, and hypernym-hyponym. Gábor et al. (2018) developed a dataset containing 500 abstracts from the computational linguistics domain and annotated various semantic relations, including part-whole, used-for, and compare. The SciERC dataset further expanded the range of annotated relation types by incorporating additional relations, such as hypernym-hyponym (Luan et al., 2018).

#### (2) Full-Text datasets for knowledge entity relationship extraction task

Full-Text datasets refer to datasets constructed from complete academic papers or selected paragraphs and sentences from full-text academic papers. For instance, Augenstein et al. (2017) constructed a dataset comprising 600 paragraphs collected from academic papers across three disciplines, including computer science, and annotated sentence-level hypernym-hyponym relations between methods and tasks. The TD-TM-Rel dataset contains 600 sentences randomly sampled from academic papers and focuses on two relation types, namely with-metric and with-dataset. These relations indicate, respectively, that a metric is used to evaluate a task and that a dataset is used to test a task (Mondal et al., 2021). Similarly, the SciNLP dataset, constructed from 60 full-text scientific papers, also includes these two types of relations (Duan et al., 2025). The SciER dataset consists of 106 full-text scientific papers and covers relations among three types of scientific entities, namely tasks, methods, and datasets. It annotates a diverse range of relations, such as used-for, part-whole, compare, synonym-of, and hypernym-hyponym (Zhang et al., 2024b). In addition, the SCIREX dataset, introduced by Jain et al. (2020), contains 438 full-text academic papers in the machine learning domain. It provides annotations for four types of entities, including datasets, metrics, methods, and tasks, as well as their associated 4-ary relations.

In KERE, LLMs have been applied in various domains, including biomedicine, clinical medicine, and materials chemistry. A range of strategies have been explored, such as prompt learning, LLM adaptation, LLM-enhanced SLMs, and SLM-enhanced closed-source LLMs. The potential of SLM-enhanced open-source LLMs and closed-source LLM-enhanced open-source LLMs for KERE remains relatively underexplored in existing studies. Exploring these strategies may provide new perspectives on KERE in academic papers. In addition, existing KERE datasets exhibit considerable heterogeneity in the definitions of relation types. Some datasets focus on relations such as part-whole, hypernym-hyponym, and compare, whereas others emphasize with-metric and with-dataset relations. Such heterogeneity poses challenges to the construction of a unified relation schema. To mitigate this issue, this study first systematically reviews the relation types defined in existing datasets through a literature review. It then further analyzes academic papers in the NLP domain to identify relation types that have not yet been adequately defined in previous studies but may have substantial value for downstream applications.

# 3. Methodology

## 3.1 Research framework

As illustrated in Fig.1, this study systematically investigates five types of LLM-based KERE models, organized around two core paradigms: closed-source LLM-based CoT prompting and LoRA-adapted open-source LLMs. These paradigms further support three complementary enhancement strategies: closed-source LLM-enhanced open-source LLMs, LLM-enhanced SLMs, and SLM-enhanced LLMs.

To analyze the performance of different types of models on the KERE task, this study constructs Abstract and Full-Text NLP datasets and refines them through error correction. To answer RQ1 and RQ2, the five LLM-based KERE strategies are evaluated using multiple metrics, and their performance advantages over SLM fine-tuning are systematically analyzed. To answer RQ3, an error analysis is conducted to identify targeted directions for further optimizing LLM-based KERE models.

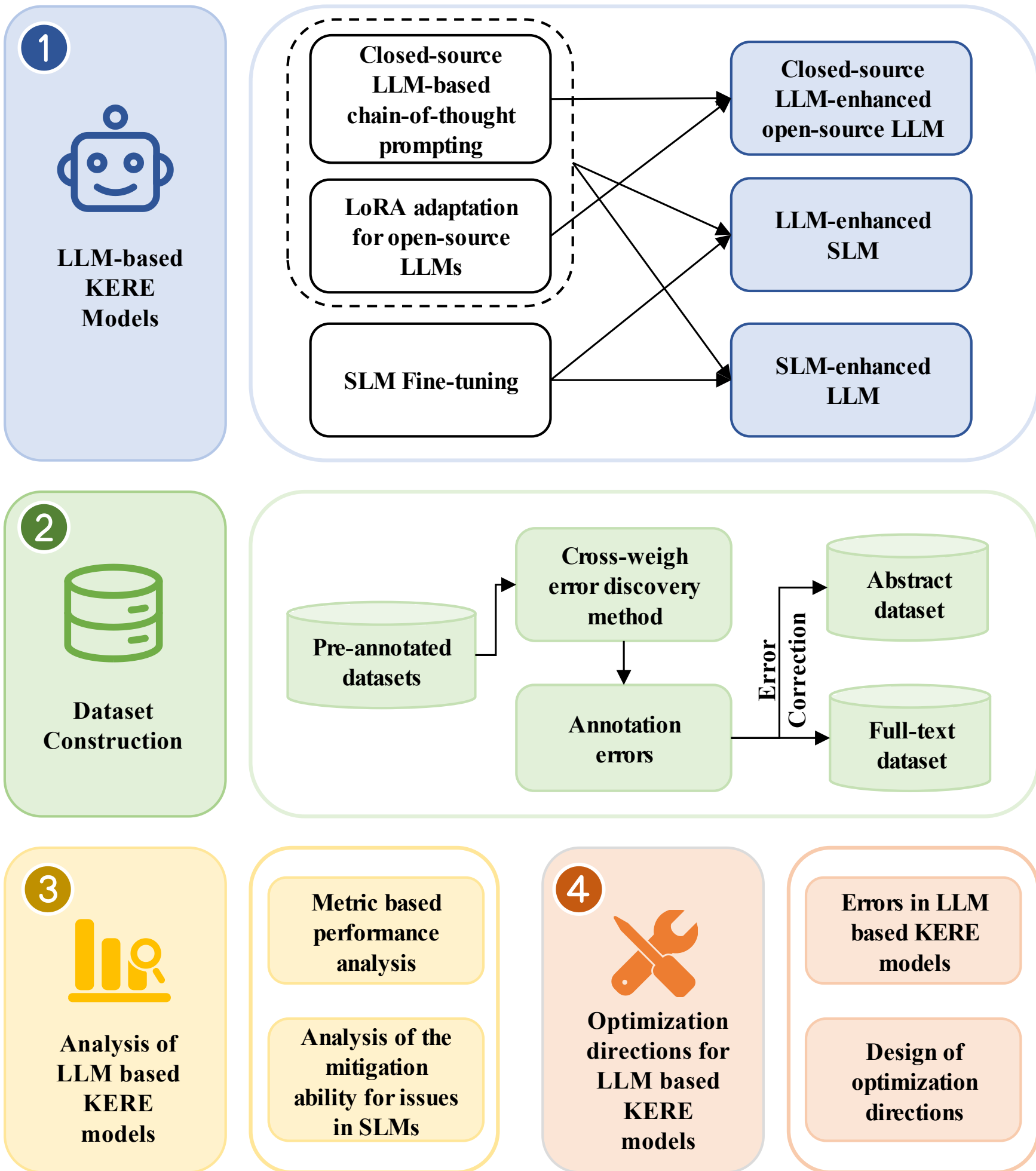

Fig 1. Research framework of this study

### 3.2 LLM based knowledge entity relationship extraction strategies

This study investigates five LLM-based strategies for KERE, which are introduced in the following subsections.

#### 3.2.1 Closed-source LLM-based Chain-of-Thought prompting

This study employs the CoT framework proposed by Ma et al. (2023). The overall architecture consists of three core components, namely a human-instructed reasoning module, a similarity-based KNN retrieval module, and an inference module.

First, the human-instructed reasoning module associates each sample in the training set with a reasoning process by prompting an LLM with manually annotated seed examples for each relation type. Second, the similarity-based KNN retrieval module selects samples of each relation type and their associated reasoning processes from the training set based on their similarity to the test sample. The selected samples are then used as few-shot demonstrations in the final prompt. Third, the inference module predicts the relation type of the test sample by prompting the LLM with the final prompt, which integrates system information, few-shot demonstrations, and user information. The overall prompt design is illustrated in Fig. 2.

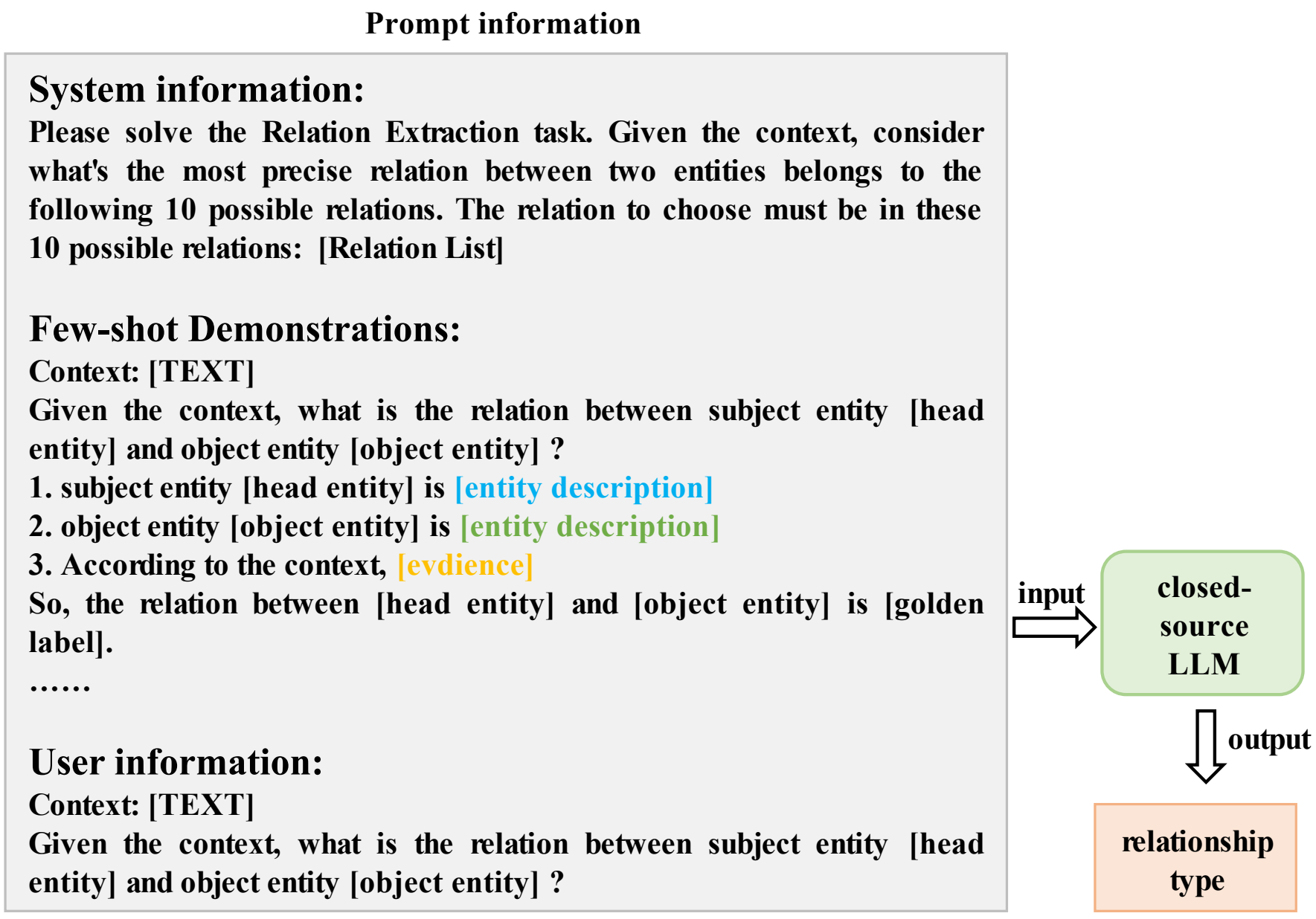

Fig 2. Flowchart of the closed-source LLM-based Chain-of-Thought prompting

### 3.2.2 LoRA-adapted open-source LLMs

Open-source LLMs generally have more parameters and are trained on larger corpora than SLMs. To adapt these models to the KERE task, this study applies LoRA-based adaptation to open-source LLMs.

During adaptation, the same system information and user information shown in Fig.2 are used. For supervised learning, the ground-truth relation labels are further incorporated into the prompts.

### 3.2.3 LLM-enhanced SLMs

To mitigate the semantic limitations of SLMs, this study uses a closed-source LLM-enhanced SLM framework in which closed-source LLMs serve as entity description generators (Asada & Fukuda, 2024).

As illustrated in Fig.3a, for each entity $e_i$, a prompt is constructed and provided to the closed-source LLM to generate a textual description $d_i$. The augmented input is then constructed as :

$$x_i' = Concat(x_i, d_i) \tag{1}$$

where $x_i$ denotes the original input containing entity $e_i$, and $x_i'$ denotes the augmented input incorporating the generated entity description. The augmented input is subsequently used by the SLM for training and testing.

### 3.2.4 SLM-enhanced LLMs

For SLM-enhanced closed-source LLMs, this study adopts an existing method that incorporates prior knowledge learned by an SLM into closed-source LLM-based CoT prompting (Tang et al., 2024). Specifically, an SLM is first trained on manually annotated data to obtain a probability distribution over relation types:

$$p = SLM(x) = [p_1, p_2, \dots, p_K], \sum_{i=1}^{K} p_i = 1 \tag{2}$$

Where $x$ denotes the input instance and $K$ denotes the number of relation types. As shown in Fig.3b, the top-N relation labels with the highest probabilities are selected and incorporated into the CoT prompt as prior guidance. This strategy reduces the candidate label space for reasoning. The augmented prompt is subsequently used by the closed-source LLM to predict the final relation type.

Building on this strategy, this study further proposes an SLM-enhanced open-source LLM method by integrating SLM-based relation type selection into the LoRA-adapted open-source LLM framework. During adaptation, the top-N relation labels predicted by the SLM are incorporated into the training prompt, allowing the open-source LLM to focus on a reduced label space. During testing, the same strategy is used to construct a candidate label set for each test sample, and the LoRA-adapted open-source LLM predicts the relation type within the reduced label space.

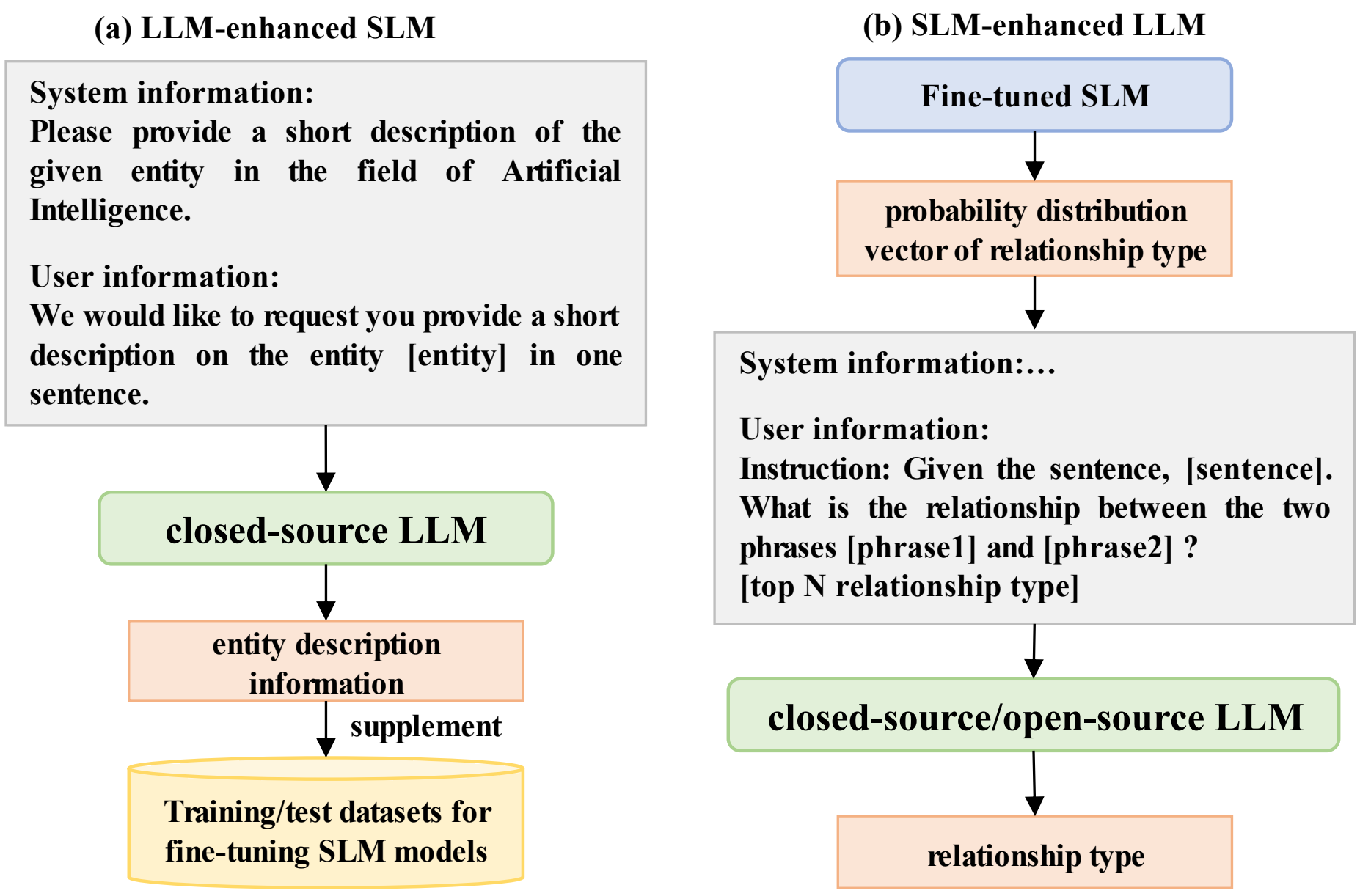

Fig 3. Flowchart of the LLM-enhanced SLMs and SLM-enhanced LLMs

### 3.2.5 Closed-source LLM-enhanced open-source LLM

To further enhance open-source LLMs, this study proposes two strategies that leverages closed-source LLMs from two perspectives, namely relation explanation generators and sample quality evaluators. First, closed-source LLMs are used to enrich the training data by generating explanations for relation results. Second, closed-source LLMs are used to evaluate and filter training samples, thereby reducing the size of the training dataset and improving adaptation efficiency.

**(1) Closed-source LLMs as relation explanation generators**

A closed-source LLM is used to generate natural language explanations for the relation results of entity pairs. As shown in Fig.4a, each prompt contains the sentence, entity pair, and ground-truth relation label. The generated explanation is combined with the relation label to construct an augmented training sample. The augmented training samples are then used for LoRA-based adaptation of the open-source LLM.

**(2) Closed-source LLMs as sample quality evaluators**

To reduce annotation noise and improve adaptation efficiency, a closed-source LLM is used to assign a quality score $s_i \in [0,5]$ to each sample $x_i$ in the training dataset. Samples with $s_i \geq T$ (where $T = 4$) are retained, while others are filtered out. The resulting high-quality subset is then used for LoRA-based adaptation of the open-source LLM. The prompt used for sample quality evaluation is illustrated in Fig.4b.

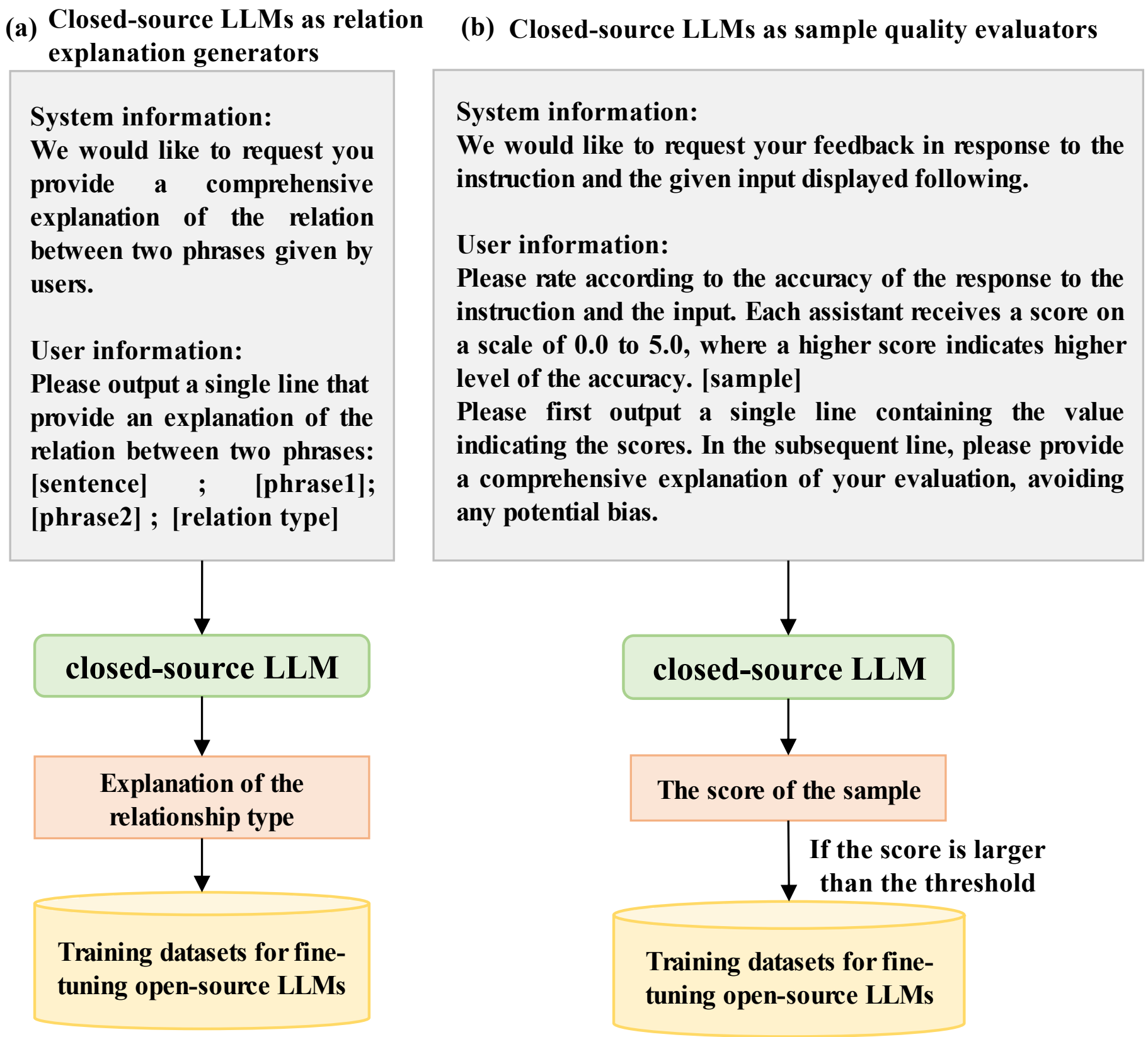

Fig 4. Flowchart of the Method for closed-source LLM-enhanced open-source LLM

# 4. Experiments and result analysis

This section presents the experimental settings and experimental results. The experimental settings introduce the LLMs and SLMs used in the experiments, model parameter configurations, dataset construction and error correction procedures, and evaluation metrics. The subsequent analysis compares the performance of five LLM-based KERE strategies, conducts ablation analyses, and examines the ethical considerations and computational costs associated with LLMs to answer RQ1. Finally, the ability of LLM-based KERE models to mitigate SLM errors is examined to answer RQ2, while potential optimization directions for LLM-based KERE are explored to answer RQ3.

## 4.1 Experimental settings

This section first introduces the open-source and closed-source LLMs selected for this study, together with the SLMs. It then describes the parameter settings for SLM training and LoRA-based adaptation of open-source LLMs. Finally, it presents the construction of the Abstract and Full-Text datasets in the NLP domain used for training and testing, as well as the evaluation metrics.

### 4.1.1 Model selection

**(1) Closed-source LLMs**

**Qwen-turbo.** Qwen-turbo is a lightweight model designed to provide cost-effective inference while maintaining competitive performance. In this study, “Qwen-turbo” model is employed.

**GPT-5 mini.** GPT-5 mini is a lightweight multimodal model designed to reduce inference costs while retaining strong model capabilities. In this study, “GPT-5 mini” is employed and accessed through the OpenAI API.

**(2) Open-source LLMs**

**T5 Model (T5).** T5 is an open-source language model developed by Google based on an encoder-decoder architecture (Raffel et al., 2020). Building on T5, the Allen Institute for AI adapted the model to extractive, abstractive, and multiple-choice question answering tasks, resulting in UnifiedQA (Khashabi et al., 2020). In this study, the unifiedqa-3B model is selected.

**LLAMA2 Model (LA).** LLAMA2 is an open-source LLM developed by Meta (Touvron et al., 2023). It is trained on a corpus of 2 trillion tokens collected from web pages, Wikipedia, ArXiv, and other sources. The model adopts a decoder-only architecture and is available in several parameter scales, including 7B, 13B, and 70B. Meta further developed LLAMA2-Chat for dialogue applications through instruction tuning and reinforcement learning from human feedback. In this study, the 7B versions of LLaMA 2 and LLaMA 2-Chat, denoted as LA and LA-C, respectively, are selected.

**ChatGLM3 Model (GL).** ChatGLM3 is a decoder-only LLM jointly developed by Zhipu AI and the KEG Laboratory of Tsinghua University (GLM et al., 2024). In this study, the open-source ChatGLM3-6B model is selected.

**Baichuan2 Model (BC).** Baichuan2 is a decoder-based LLM developed by Baichuan Intelligence (Yang et al., 2023). It is trained on a corpus comprising 2.6 trillion characters and is available in 7B and 13B parameter versions. Baichuan2-Chat is a variant of Baichuan2 designed for dialogue tasks. In this study, the 7B version of Baichuan2-Chat is selected.

**Qwen 2.5 Model (QW).** Qwen 2.5 is an open-source LLM released by Alibaba DAMO Academy (Qwen et al., 2025). It adopts a decoder-only architecture. In this study, Qwen2.5-7B is selected.

**DeepSeek-R1-Distill-Llama model（D-L）**. The DeepSeek-R1-Distill-Llama model is an open-source LLM released by DeepSeek and developed through knowledge distillation based on Meta's LLaMA architecture (DeepSeek-AI et al., 2025). It adopts a decoder-only architecture. In this study, DeepSeek-R1-Distill-Llama-8B is selected.

**(3) SLM-based relation extraction models**

**R-BERT model (R-BERT).** R-BERT consists of three main components, including a sentence representation module, an entity representation module, and a relation classification module (Wu & He, 2019). BERT is used to encode the input sentence and target entities. The sentence-level and entity-level representations are then concatenated and provided to the relation classification module, which predicts the probability distribution over the predefined relation types.

**Longformer based model (Longformer).** The Longformer-based model consists of an embedding module, a feature extraction module, and a relation classification module. The embedding module uses Longformer (Beltagy et al., 2020) to generate contextualized representations of the input text and capture long-range dependencies. The feature extraction module then combines BiLSTM and Transformer architectures to extract semantic features relevant to the KERE task. Finally, the relation classification module predicts the probability distribution over the predefined relation types.

**RE-SciBERT model (RE-SciBERT).** RE-SciBERT first inserts special marker tokens around the target entities to explicitly indicate their positions in the input text (Aid et al., 2024). The marked sequence is then encoded by SciBERT (Beltagy et al., 2019) to generate contextualized representations. The representations of the target entities and the global sentence representation are subsequently extracted and concatenated as input to a multi-layer neural network for relation classification.

#### 4.1.2 Parameter settings

**(1) LLM parameter settings**

The open-source LLMs are adapted using an NVIDIA GeForce RTX 4090 GPU with 24 GB of VRAM. Considering the memory requirements of different models, LoRA is applied for model adaptation. Two open-source Python libraries are used to implement the adaptation process. PEFT provides the implementation of LoRA, while TRL supports supervised training. During adaptation, the number of epochs is set to 2, the gradient accumulation step is set to 4, and the learning rate is set to 3e-4.

**(2) SLM-based model parameter settings**

BERT, Longformer, and SciBERT are implemented using the Transformers library, version 3.0.2. The pretrained models bert-base-uncased, longformer-base-4096, and scibert-scivocab-uncased are selected, respectively. The number of training epochs is set to 20 for the Abstract dataset and 10 for the Full-Text dataset. The batch size is set to 16, and the Adam optimizer is used with a learning rate of 3e-5. The maximum token length is set to 512. For feature extraction, the BiLSTM layer contains 150 hidden units, and the self-attention mechanism is configured with 8 attention heads.

#### 4.1.3 Dataset construction

**(1) Preparation of datasets for annotation**

**Abstract Dataset.** This study uses the publicly available SciERC dataset proposed by Luan et al. (2018), which contains 500 abstracts collected from 12 major artificial intelligence conferences covering NLP and machine learning. From the SciERC dataset, 305 abstracts related to the NLP domain are selected for annotation.

**Full-Text Dataset.** Full-Text academic papers are collected from the ACL Anthology Corpus, which contains XML-formatted papers published in the proceedings of the Annual Meeting of the Association for Computational Linguistics from 1979 to 2015, comprising a total of 4,641 papers[1]. To construct the Full-Text dataset, two papers are randomly sampled from each year, resulting in a total of 74 full-text papers.

**(2) Construction of the relation type scheme in the NLP domain**

This study defines relation types in the NLP domain based on both a literature review and data analysis. Based on the literature review, six existing relation types are adopted, including used-for, hypernym-hyponym represented as superclass-of, part-whole represented as part-of, compare, with-metric, and with-dataset. Through further analysis of academic papers in the NLP domain, three additional relation types are identified, namely sequence-of, directed-compare, and problem-exist. Examples of these relation types from academic papers are presented in

---

[1] The URL of the ACL ARC corpus is https://acl-arc.comp.nus.edu.sg/ (this URL is currently invalid).

Table 1.

The sequence-of relation captures the sequential order of two tasks. For example, in Table 1, "ranks the list of output sentence plans" precedes "selects the top-ranked plan" and serves as a prerequisite step. This relation can support procedural step extraction and workflow construction (Gu et al., 2025). The directed-compare relation represents a directional comparison between two entities. For example, the relation between "lemma-based model" and "word form model" indicates that one method outperforms the other. Unlike the undirected compare relation, directed-compare is useful for the identification of technological evolution paths (Xue & Shao, 2024). Finally, the problem-exist relation describes a problem associated with a method. For instance, "language model" is associated with "insufficient training data" in Table 1. This relation can support method limitation analysis and the identification of potential research problems (Ott., 2024 ).

Table 1. Relation types and corresponding references

| Relation Type | Explanation | References | Example |
|---|---|---|---|
| used-for | Entities are linked by usage relation | Luan et al. (2018); Augenstein et al. (2017) | An efficient **<u>bit-vector-based CKY-style parser</u>** for **<u>context-free parsing</u>** is presented. |
| superclass-of | Entities are in a hypernym-hyponym relation | Luan et al. (2018); Augenstein et al. (2017) | A method of sense resolution is proposed that is based on **<u>WordNet</u>**, an **<u>on-line lexical database</u>** that incorporates semantic relations as labeled pointers between word senses. |
| part-of | Entities are in a part-whole relation | Gábor et al. (2018); Luan et al. (2018) | The **<u>answering agents</u>** adopt fundamentally different strategies, one utilizing primarily knowledge-based mechanisms and the other adopting **<u>statistical techniques</u>**. |
| sequence-of | the sequential relationship between the execution of two tasks | - | Second, the sentence-plan-ranker (SPR) **<u>ranks the list of output sentence plans</u>**, and then **<u>selects the top-ranked plan</u>**. |
| with-metric | Tasks are evaluated with metrics | Mondal et al. (2021); Duan et al. (2025) | The **<u>accuracy rate</u>** of **<u>syntactic disambiguation</u>** is raised from 46.0% to 60.62% by using this novel approach. |
| compare | Entities are linked by comparison relation | Gábor et al. (2018); Luan et al. (2018) | Further, in their optimum configuration, **<u>bag-of-words methods</u>** are shown to be equivalent to **<u>segment order-sensitive methods</u>** in terms of retrieval accuracy, but much faster. |
| directed-compare | The two entities are linked by a directional comparison in which the first method outperforms the second, or the first task is more difficult than the second. | - | Testing the **<u>lemma-based model</u>** on the Dutch Senseval-2 test data, we achieve a significant increase in accuracy over the **<u>word form model</u>**. |
| with-dataset | Tasks are tested with datasets | Mondal et al. (2021); Duan et al. (2025) | However, our experience with **<u>TACITUS</u>**; especially in the MUC-3 evaluation, has shown that principled techniques for **<u>syntactic and pragmatic analysis</u>** can be bolstered with methods for achieving robustness. |
| problem-exist | The method suffers from certain problems. | - | Owing to the problem of **<u>insufficient training data</u>** and approximation error introduced by the **<u>language model</u>**, traditional statistical approaches , ... |
| irrelevant | There is no such relationship as mentioned above between entities | - | - |

**Note.** Bold and underlined expressions in the examples indicate the entity mentions involved in the corresponding relation. A dash (-) in the References column indicates that the relation type was introduced in this study rather than adopted from previous studies.

**(3) Dataset annotation and error correction**

After defining the relation types, relations between entities are annotated. Two annotators with academic backgrounds in NLP, including one master's student and one doctoral student, participated in a preliminary annotation process. Thirty abstracts were randomly selected from the Abstract dataset and independently annotated by the two annotators. Cohen's kappa was used to assess inter-annotator agreement, resulting in a kappa coefficient of 0.723. Given this level of agreement, the doctoral student subsequently completed the remaining annotations independently.

Table 2. Distribution of annotation and prediction errors across the two datasets

| Category / Dataset | annotation errors | | prediction errors | |
|---|---|---|---|---|
| | num | percentage | num | percentage |
| **Abstract** | 114 | 14.32 | 682 | 85.68 |
| **Full-Text** | 395 | 18.27 | 1767 | 81.73 |

As annotation errors are difficult to avoid completely, this study adopts an error detection procedure based on CrossWeigh (Wang et al., 2019) to improve dataset quality. The error detection and correction procedure in this study consists of three steps. First, ten-fold cross-validation is performed, and incorrectly predicted samples are collected to form a candidate error set. Second, the master's student annotator manually examines each sample in the candidate error set and classifies it into one of two categories, namely prediction errors, where samples are correctly annotated but incorrectly predicted, and annotation errors, where samples are incorrectly annotated but correctly predicted. Third, prediction errors are retained without modification, whereas annotation errors are reviewed with the doctoral student annotator to determine the final relation labels. The statistics of prediction and annotation errors are presented in Table 2, while the statistics of the datasets after error correction are reported in Table 3.

Table 3. Statistics of each relation type in the two datasets

| Relation type | Abstract | Full-Text |
|---|---|---|
| used-for | 773 | 2,573 |
| superclass-of | 165 | 571 |
| part-of | 363 | 1,066 |
| sequence-of | 54 | 200 |
| with-metric | 78 | 133 |
| compare | 25 | 142 |
| directed-compare | 38 | 122 |
| with-dataset | 23 | 16 |
| problem-exist | 8 | 39 |
| irrelevant | 14,048 | 42,561 |
| Total | 15,575 | 47,423 |

To support subsequent error analysis, the causes of incorrect predictions for samples in the first category are also annotated. Based on the manual analysis, three types of prediction errors are identified, including coreference resolution errors, relation direction prediction errors, and entity contextual semantic understanding errors. Section 4.3 further analyzes these three error types and examines whether LLMs can mitigate such errors in KERE.

**(4) Strategies for handling imbalanced data**

The distribution of relationship types in the dataset is imbalanced, with the irrelevant category accounting for a large proportion of the samples. Such imbalance may reduce the model's ability to identify relation types with relatively few training samples. To mitigate this imbalance, this study applies a data replication strategy to the training datasets.

Specifically, let $N$ denote the number of samples in the irrelevant category and $M$ denote the number of samples in another relation type. A divisor $t$ is introduced to control the degree of data replication, and the replication factor is calculated as $\frac{N}{tM}$. In this study, $t$ is set to 3.

#### 4.1.4 Performance evaluation methods

This study uses five-fold cross-validation to evaluate model performance. The evaluation metrics include the $F_1$ score, macro precision (P), macro recall (R), macro $F_1$, macro $F_1$ excluding the irrelevant relation type, and micro $F_1$. Section 4.3 further analyzes the extent to which LLMs can mitigate the limitations of SLMs, and section 4.4 identifies potential optimization directions for LLM-based KERE models.

### 4.2 Analysis of experimental results

To answer RQ1, this section compares the performance of different LLM-based KERE approaches. To control experimental costs, approaches involving GPT-5 mini are evaluated only on the Abstract dataset.

#### 4.2.1 Performance analysis based on evaluation metrics

**(1) Performance of closed-source LLM-based Chain-of-Thought prompting**

Table 4. Results of closed-source LLM-based CoT prompting on Abstract and Full-Text datasets

| Metrics (%) \ Model | | Abstract | | Full-Text |
|---|---|---|---|---|
| | | Qwen-turbo | GPT-5 mini | Qwen-turbo |
| $F_1$ | used-for | **29.80** | 26.90 | 31.68 |
| | superclass-of | 28.32 | **39.64** | 24.79 |
| | part-of | 21.23 | **23.10** | 21.27 |
| | sequence-of | 3.49 | **5.54** | 7.14 |
| | with-metric | **17.75** | 16.79 | 9.76 |
| | compare | 2.29 | **12.93** | 3.79 |
| | directed-compare | 18.17 | **20.74** | 17.20 |
| | with-dataset | 3.72 | **7.42** | 1.31 |
| | problem-exist | **11.88** | 10.18 | 13.75 |
| | irrelevant | 65.65 | **70.22** | 65.05 |
| Macro P | | 19.20±0.68 | **20.59**±1.46 | 18.77±0.62 |
| Macro R | | 49.44±7.91 | **63.83**±3.27 | 47.66±3.57 |
| Macro $F_1$ | | 20.61±1.16 | **23.83**±1.89 | 19.58±0.69 |
| Macro $F_1$ (w/o irrelevant) | | 15.18±2.05 | **18.14**±2.30 | 14.52±0.76 |
| Micro $F_1$ | | 53.92±0.83 | **56.72**±0.30 | 53.60±0.57 |

**Note.** Bold values indicate the best performance among the models compared for the corresponding metric on the Abstract dataset. Boldface is not applied to the Full-Text results because only Qwen-turbo was evaluated on this dataset.

**GPT-5 mini achieves better performance than Qwen-turbo on the Abstract dataset using CoT prompting.** Specifically, as shown in Table 4, GPT-5 mini improves macro $F_1$ from 0.2061 to 0.2383 and micro $F_1$ from 0.5392 to 0.5672. Its higher macro $F_1$ excluding the irrelevant relation type also suggests an advantage in identifying relation types with fewer samples.

**Relation extraction on the Full-Text dataset is more challenging than that on the Abstract dataset when**

**using CoT prompting.** As shown in Table 4, for Qwen-turbo, macro $F_1$ decreases from 0.2061 on the Abstract dataset to 0.1958 on the Full-Text dataset. This performance degradation suggests that the redundant contextual information in full-text documents may introduce additional noise and increase the difficulty of relation identification.

**(2) Performance of LoRA-adapted open-source LLMs**

Table 5. Results of LoRA-adapted open-source LLMs on Abstract dataset

| Metrics(%) \ Model | | T5 | LA | LA-C | BC | CG | QW | D-LA |
|---|---|---|---|---|---|---|---|---|
| $F_1$ | used-for | **85.31** | 84.79 | 79.10 | 82.77 | 73.23 | 67.09 | 83.00 |
| | superclass-of | **73.80** | 70.67 | 48.75 | 65.91 | 51.30 | 48.86 | 70.20 |
| | part-of | **76.21** | 71.20 | 62.34 | 71.76 | 61.15 | 56.05 | 74.65 |
| | sequence-of | **67.26** | 65.79 | 61.87 | 63.86 | 59.06 | 50.04 | 59.28 |
| | with-metric | **71.55** | 65.44 | 50.11 | 66.06 | 54.15 | 37.88 | 58.93 |
| | compare | **76.59** | 74.53 | 64.55 | 71.44 | 0.0 | 10.67 | 54.24 |
| | directed-compare | 60.35 | 53.23 | 47.78 | 59.07 | 26.46 | 22.67 | **62.95** |
| | with-dataset | 46.74 | 46.57 | 33.30 | 41.53 | **50.56** | 41.62 | 46.43 |
| | problem-exist | **87.50** | 25.00 | 22.50 | **87.50** | 0.0 | 25.00 | 75.00 |
| | irrelevant | **97.94** | 97.86 | 97.18 | 97.84 | 96.94 | 96.26 | 97.78 |
| Macro P | | 75.24±5.57 | 77.38±8.03 | 68.93±7.20 | **83.31**±3.35 | 60.37±8.19 | 58.80±6.53 | 80.60±4.93 |
| Macro R | | **75.08**±4.94 | 60.94±7.22 | 52.01±5.83 | 64.99±8.05 | 43.32±4.36 | 42.73±7.02 | 62.51±7.51 |
| Macro $F_1$ | | **74.08**±3.30 | 66.47±6.14 | 57.50±5.40 | 70.49±4.83 | 48.33±4.81 | 46.11±3.22 | 68.07±6.11 |
| Macro $F_1$ (w/o irrelevant) | | **71.37**±3.61 | 62.91±6.74 | 53.91±6.73 | 67.39±5.31 | 42.92±5.34 | 40.41±3.38 | 64.68±6.80 |
| Micro F1 | | **96.15**±0.50 | 96.02±0.51 | 94.74±0.48 | 95.95±0.35 | 94.24±0.59 | 93.04±0.67 | 95.88±0.51 |

**Note.** Bold values indicate the best performance among the models compared for the corresponding metric on the Abstract dataset.

Table 6. Results of LoRA-adapted open-source LLMs on Full-Text dataset

| Metrics(%) \ Model | | T5 | LA | LA-C | BC | CG | QW | D-LA |
|---|---|---|---|---|---|---|---|---|
| $F_1$ | used-for | **88.22** | 87.27 | 86.75 | 75.64 | 83.34 | 83.36 | 68.98 |
| | superclass-of | 73.32 | **74.22** | 75.33 | 52.23 | 71.19 | 68.30 | 36.37 |
| | part-of | **73.89** | 73.48 | 74.35 | 57.96 | 69.60 | 69.17 | 45.73 |
| | sequence-of | **77.09** | 75.43 | 74.30 | 58.43 | 64.99 | 72.66 | 46.24 |
| | with-metric | **61.41** | 58.30 | 60.73 | 53.73 | 59.32 | 52.26 | 40.68 |
| | compare | 77.23 | **78.76** | 72.69 | 75.07 | 2.42 | 66.24 | 63.66 |
| | directed-compare | 56.93 | 51.85 | **58.53** | 48.07 | 36.07 | 45.20 | 34.80 |
| | with-dataset | **34.02** | 10.00 | 15.00 | 30.52 | 24.76 | 5.00 | 16.89 |
| | problem-exist | **61.51** | 51.64 | 47.53 | 36.16 | 37.15 | 39.66 | 25.63 |
| | irrelevant | 97.66 | **97.97** | **97.97** | 95.20 | 97.32 | 97.47 | 92.41 |
| Macro P | | 64.72±2.63 | 74.12±2.09 | **75.02**±4.94 | 60.40±8.91 | 55.33±5.19 | 69.71±3.46 | 39.91±7.85 |
| Macro R | | **79.45**±4.29 | 60.95±2.02 | 61.36±3.28 | 61.39±4.32 | 58.16±4.89 | 54.84±1.87 | 65.63±3.90 |
| Macro $F_1$ | | **70.13**±2.32 | 65.89±1.50 | 66.32±2.96 | 58.30±8.78 | 54.62±3.85 | 59.93±1.90 | 47.14±6.63 |
| Macro $F_1$ (w/o irrelevant) | | **67.07**±2.57 | 62.33±1.66 | 62.80±3.29 | 52.42±7.94 | 49.87±4.25 | 55.76±2.11 | 42.11±7.05 |
| Micro $F_1$ | | 95.74±0.13 | **96.27**±0.11 | **96.27**±0.16 | 91.45±0.39 | 94.91±0.19 | 95.36±0.17 | 86.84±5.07 |

**Note.** Bold values indicate the best performance among the models compared for the corresponding metric on the Full-Text dataset.

**T5 achieves the highest macro $F_1$ scores on both the Abstract and Full-Text datasets using LoRA adaptation.** As shown in Table 5, on the Abstract dataset, T5 achieves a macro $F_1$ of 0.7408 and a macro $F_1$ excluding the irrelevant relation type of 0.7137, together with a macro precision of 0.7524 and a macro recall of 0.7508. Similarly, as shown in Table 6, on the Full-Text dataset, T5 achieves a macro $F_1$ of 0.7013 and a macro $F_1$ excluding the irrelevant relation type of 0.6707. These results demonstrate the relatively consistent performance of T5 across the two datasets.

**Different models exhibit distinct strengths across evaluation metrics.** As shown in Table 5, on the Abstract dataset, Baichuan2 achieves the highest macro precision of 0.8331, substantially exceeding the 0.7524 achieved by T5. However, as shown in Table 6, this advantage is not observed on the Full-Text dataset, where LLaMA2-Chat achieves the highest macro precision of 0.7502, compared with 0.6472 for T5. These results suggest that the model with the strongest precision performance may vary across datasets.

**(3) Performance of LLM-enhanced SLMs**

Table 7. Results of SLMs on Abstract and Full-Text datasets

| Metrics(%) \ Model | | Abstract | | | Full-Text | | |
|---|---|---|---|---|---|---|---|
| | | RE-SciBERT | Longformer | R-BERT | RE-SciBERT | Longformer | R-BERT |
| F1 | used-for | **80.67** | 77.91 | 79.41 | **86.43** | 83.26 | 85.48 |
| | superclass-of | **64.56** | 61.16 | 57.32 | **66.81** | 64.58 | 65.98 |
| | part-of | **65.51** | 65.36 | 63.74 | 71.65 | 68.46 | **71.86** |
| | sequence-of | 46.52 | **55.60** | 46.72 | **65.74** | 65.24 | 63.64 |
| | with-metric | **69.82** | 65.20 | 62.43 | **64.54** | 56.62 | 63.62 |
| | compare | 79.44 | 65.54 | **87.62** | 87.69 | 77.62 | **89.84** |
| | directed-compare | **42.63** | 40.00 | 26.98 | 54.61 | **57.02** | 53.56 |
| | with-dataset | **54.36** | 49.69 | 43.00 | 25.55 | **32.63** | 13.40 |
| | problem-exist | **75.00** | 66.75 | 50.00 | **54.20** | 43.80 | 52.52 |
| | irrelevant | **97.48** | 97.07 | 97.25 | 97.81 | 97.00 | **97.96** |
| Macro P | | **78.11**±7.13 | 66.65±5.75 | 71.07±10.97 | 73.31±4.05 | 62.21±4.76 | **73.72**±4.71 |
| Macro R | | 62.32±4.86 | **67.70**±4.69 | 57.32±8.13 | 64.60±4.59 | **71.79**±6.38 | 61.06±2.92 |
| Macro $F_1$ | | **67.40**±4.22 | 64.35±3.90 | 61.68±7.93 | **67.50**±3.59 | 64.47±4.28 | 65.34±3.51 |
| Macro $F_1$ (w/o irrelevant) | | **63.96**±4.74 | 60.67±4.36 | 57.63±8.80 | **64.13**±3.99 | 60.96±4.76 | 62.14±3.69 |
| Micro $F_1$ | | **95.27**±0.35 | 94.41±0.56 | 94.82±0.43 | 95.92±0.12 | 94.68±0.55 | **96.04**±0.23 |

**Note.** Bold values indicate the best performance among the models compared for the corresponding metric on the Abstract dataset and the Full-Text dataset.

To identify a strong-performing SLM for subsequent LLM-enhanced SLM experiments, three representative SLM-based relation extraction models are evaluated on both Abstract and Full-Text datasets. As shown in Table 7, on the Abstract dataset, RE-SciBERT achieves the highest macro $F_1$ of 0.6740, followed by Longformer and R-BERT with 0.6435 and 0.6168. A similar trend is observed on the Full-Text dataset, where RE-SciBERT and R-BERT achieve macro $F_1$ scores of 0.6750 and 0.6534, respectively. Based on its consistently strong macro $F_1$ performance, RE-SciBERT is selected as the SLM for subsequent experiments.

For the LLM-enhanced SLM strategy, RE-SciBERT is selected as the SLM and GPT-5 mini as the closed-source LLM. **As shown in Table 8, semantic augmentation using LLM-generated entity descriptions improves the performance of RE-SciBERT on both datasets.** Specifically, GPT-5 mini-enhanced RE-SciBERT achieves macro $F_1$ scores of 0.6844 on the Abstract dataset and 0.6778 on the Full-Text dataset, compared with 0.6740 and 0.6750, respectively, for the original RE-SciBERT model. To further examine the robustness of this strategy, R-BERT was additionally evaluated on the Abstract dataset. The results show that GPT-5 mini-enhanced R-BERT achieves a macro $F_1$ score of 0.6320, compared with 0.6168 for the original R-BERT model, representing an absolute improvement of 0.0152. These results suggest that incorporating LLM-generated semantic information can provide additional benefits for SLM-based KERE. Meanwhile, LLM-generated entity descriptions may contain hallucinated information that may affect the performance of relation extraction. The impact of such errors is further examined through the ablation analysis in Section 4.2.2.

Table 8. Results of LLM-enhanced SLMs on Abstract and Full-Text datasets

| Metrics(%) \ Model | | Abstract | Full-Text |
|---|---|---|---|
| | | GPT-5 mini enhanced RE-SciBERT | GPT-5 mini enhanced RE-SciBERT |
| $F_1$ | used-for | 80.29 | 85.94 |
| | superclass-of | 61.10 | **66.98** |
| | part-of | 64.77 | **71.96** |
| | sequence-of | **56.13** | 62.90 |
| | with-metric | 66.08 | **68.96** |
| | compare | **84.94** | 87.38 |
| | directed-compare | **48.74** | 52.17 |
| | with-dataset | 53.33 | **28.89** |
| | problem-exist | 72.50 | **54.85** |
| | irrelevant | 97.38 | 97.78 |
| Macro P | | 76.15±7.68 | **74.00**±7.08 |
| Macro R | | **64.92**±5.25 | **64.92**±4.59 |
| Macro $F_1$ | | **68.44**±5.35 | **67.78**±4.75 |
| Macro $F_1$ (w/o irrelevant) | | **65.15**±5.92 | **64.45**±5.28 |
| Micro $F_1$ | | 95.12±0.58 | 95.88±0.12 |

**Note.** Bold values indicate that GPT-5 mini enhanced RE-SciBERT achieves a numerically higher result than the original RE-SciBERT. The original RE-SciBERT results are reported in Table 7.

**(4) Performance of SLM-enhanced LLMs**

For SLM-enhanced LLM strategies, RE-SciBERT is selected as the SLM, while GPT-5 mini and T5 are selected as the closed-source and open-source LLMs, respectively. The experimental results are presented in Table 9.

Table 9. Results of SLM-enhanced LLMs on Abstract and Full-Text datasets

| Metrics(%) \ Model | | Abstract | | Full-Text |
|---|---|---|---|---|
| | | RE-SciBERT enhanced LLMs | | RE-SciBERT enhanced T5 |
| | | GPT-5 mini | T5 | |
| $F_1$ | used-for | 32.65 | 85.00 | 85.38 |
| | superclass-of | 32.39 | 73.01 | 61.36 |
| | part-of | 31.87 | 75.15 | 64.10 |
| | sequence-of | 10.86 | 67.86 | 68.21 |
| | with-metric | 30.73 | 72.74 | 49.42 |
| | compare | 24.05 | 76.05 | 28.79 |
| | directed-compare | 21.55 | 54.72 | 57.24 |
| | with-dataset | 8.59 | 45.07 | 17.33 |
| | problem-exist | 19.57 | 62.50 | 45.16 |
| | irrelevant | 77.42 | 97.90 | 97.43 |
| Macro P | | 25.65±5.63 | 72.94±10.45 | 64.99±1.87 |
| Macro R | | 58.47±6.24 | 71.47±7.91 | 53.47±3.70 |
| Macro $F_1$ | | 29.71±4.22 | 71.26±8.61 | 57.48±2.10 |
| Macro $F_1$ (w/o irrelevant) | | 23.58±2.96 | 68.24±9.50 | 53.00±3.52 |
| Micro $F_1$ | | 66.72±1.10 | 96.08±0.43 | 95.12±0.58 |

**The SLM-enhanced open-source LLM performs worse than the corresponding LoRA-adapted open-source LLM on both datasets.** Specifically, the top-3 relation labels predicted by RE-SciBERT are incorporated into T5 as additional candidate information. RE-SciBERT-enhanced T5 achieves macro $F_1$ scores of 0.7126 on the Abstract dataset and 0.5748 on the Full-Text dataset, both lower than those achieved by the LoRA-adapted T5 model. The effect of the number of candidate relation labels is further examined in the ablation analysis in Section 4.2.2.

**In contrast, the SLM-enhanced closed-source LLM improves the performance of GPT-5 mini-based**

**CoT prompting.** When the top-3 relation labels predicted by RE-SciBERT are incorporated into prompt, the macro $F_1$ score increases from 0.2383 to 0.2971 on the Abstract dataset. This result suggests that SLM-generated candidate relations can provide useful guidance for closed-source LLMs. Nevertheless, the resulting performance remains substantially lower than that of LoRA-adapted open-source LLMs.

**(5) Performance of the closed-source LLM enhanced open-source LLM**

This study selects T5 as the representative open-source LLM. Qwen2.5 is additionally selected as a representative decoder-only model to examine whether the effects of the proposed strategies vary across model architectures. GPT-5 mini is selected as the closed-source LLM.

**Incorporating relation explanations generated by the closed-source LLM into the training data does not improve the performance of T5, whereas it benefits the decoder-only Qwen2.5 model on the Abstract dataset.** As shown in Table 10, the $F_1$ score of T5 decreases when relation explanations are incorporated into the training data. In contrast, Qwen2.5 achieves an $F_1$ score improvement of 0.151 under the same strategy. These results suggest that the effectiveness of explanation-based data augmentation may vary across model architectures.

**Using a closed-source LLM to select high-quality training samples also does not lead to performance improvements.** As shown in Table 10, the performance of open-source LLMs adapted using the filtered training data decreases. To further investigate this result, in Section 4.2.2, this study analyzes the influence of the sample selection threshold and compares the LLM-based sample selection strategy with random sample selection.

Table 10. Results of closed-source LLM enhanced open-source LLM methods

| Model / Metrics (%) | | **Abstract dataset** | | | **Full-Text dataset** | | |
|---|---|---|---|---|---|---|---|
| | | closed-source LLM as the result explanation tool enhanced | | closed-source LLM as the sample quality evaluation tool enhanced T5 | closed-source LLM as the result explanation tool enhanced | | closed-source LLM as the sample quality evaluation tool enhanced T5 |
| | | T5 | QW | | T5 | QW | |
| $F_1$ | used-for | 70.93 | **69.69** | 78.50 | 79.92 | 78.19 | 85.17 |
| | superclass-of | 59.84 | 45.10 | 63.61 | 61.97 | 58.99 | 72.26 |
| | part-of | 47.59 | 49.82 | 62.66 | 53.05 | 58.90 | 69.21 |
| | sequence-of | 49.88 | 45.98 | 53.93 | 56.42 | 61.07 | 72.31 |
| | with-metric | 50.77 | **56.51** | 54.33 | 52.51 | **53.22** | 52.79 |
| | compare | 47.63 | **40.43** | 54.76 | 54.70 | 62.58 | 62.93 |
| | directed-compare | 41.43 | **31.95** | 53.29 | 43.97 | **46.71** | 55.50 |
| | with-dataset | 32.23 | 23.79 | 28.25 | **45.65** | **27.67** | 33.97 |
| | problem-exist | 36.67 | 10.00 | 72.50 | 34.85 | 34.92 | 60.19 |
| | irrelevant | 93.70 | 94.78 | 95.59 | 94.81 | 96.00 | 96.91 |
| Macro P | | 43.74±1.12 | 46.35±2.51 | 53.00±2.71 | 47.06±1.83 | 55.03±4.73 | 56.60±3.30 |
| Macro R | | **78.03**±5.48 | **52.98**±6.13 | **83.64**±2.55 | **84.04**±2.65 | **64.32**±6.50 | **85.08**±3.51 |
| Macro $F_1$ | | 53.45±1.63 | **47.62**±2.33 | 61.51±1.22 | 57.79±1.62 | 57.82±5.13 | 66.12±2.97 |
| Macro $F_1$ (w/o irrelevant) | | 48.87±1.60 | **42.26**±2.41 | 57.62±1.42 | 53.67±1.76 | 53.58±5.69 | 62.70±3.27 |
| Micro $F_1$ | | 88.89±0.76 | 90.62±0.59 | 92.12±0.54 | 90.81±0.76 | 92.79±0.32 | 94.41±0.40 |

**Note.** Bold values indicate that the corresponding closed-source LLM enhanced open-source LLM methods achieves a numerically higher result than its LoRA-adapted open-source LLM baselines on the same dataset and evaluation metric. Results for the LoRA-adapted open-source LLM baselines are reported in Tables 5 and 6.

### 4.2.2 Ablation analysis

**(1) Analysis of the influence of hallucinated entity descriptions in LLM-enhanced SLMs**

To investigate the robustness of the model against hallucinated entity definitions, this study introduces controlled noise by randomly replacing a proportion of LLM-generated entity explanations at different ratios on the abstract dataset.

**As shown in Fig. 5a, model performance gradually declines as the proportion of hallucinated entity descriptions increases.** Under the clean setting, the model achieves a macro $F_1$ of 0.6844. When 5% of the entity descriptions are replaced, the macro $F_1$ slightly decreases to 0.6826. As the replacement ratio increases to 10% and 20%, the macro $F_1$ further decreases to 0.6556 and 0.6523, respectively. Despite this downward trend, the model maintains relatively consistent performance across different hallucination levels. These results suggest that the framework does not rely heavily on each individual entity description and can tolerate a certain degree of hallucination in LLM-generated semantic information.

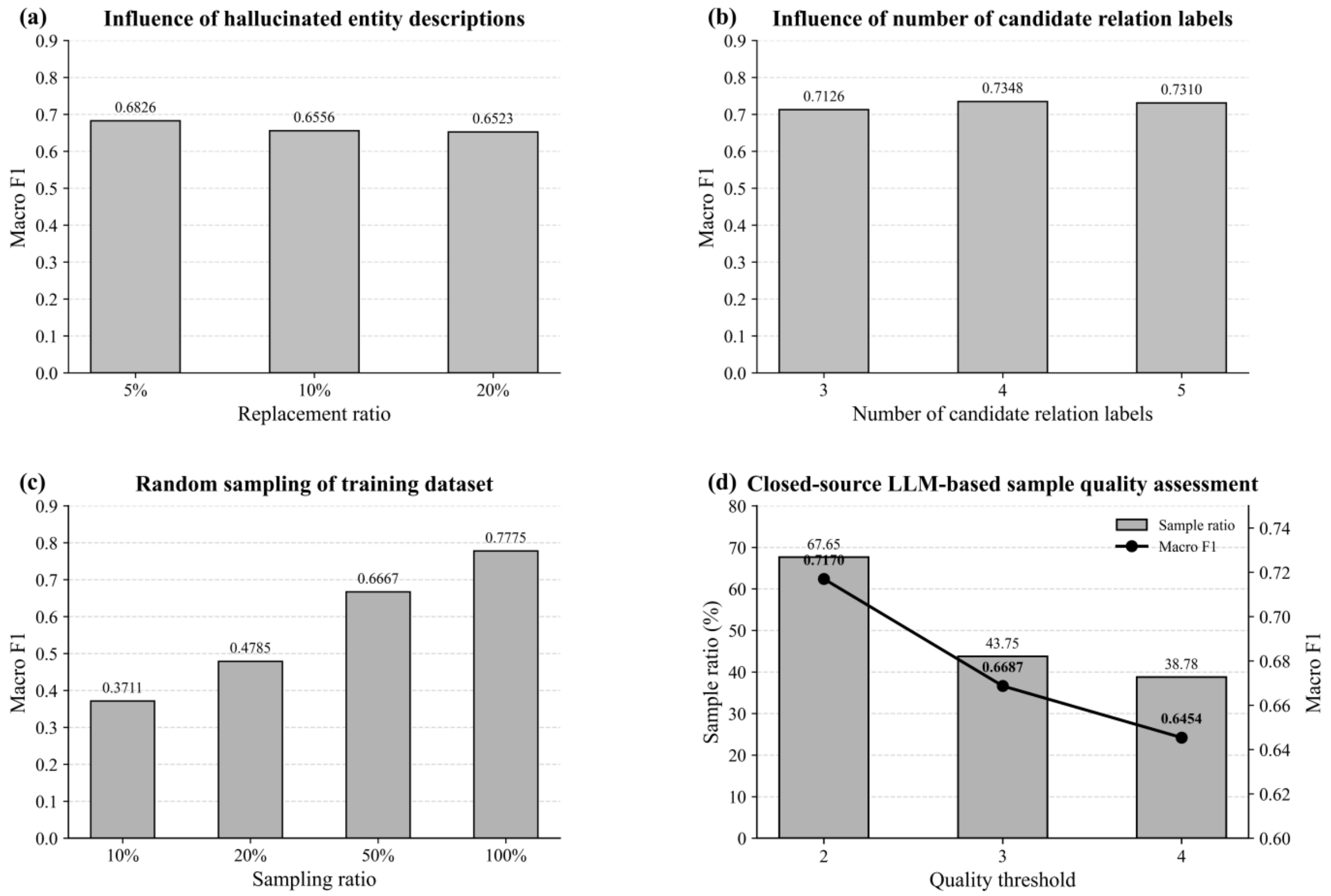

Fig 5. The results of ablation analysis

**(2) Analysis of the number of candidate relation labels for SLM-enhanced LLMs**

To investigate the impact of the number of selected relation labels on model performance, T5 is evaluated under different top-k settings on the Abstract dataset.

**As shown in Fig. 5b, increasing the number of candidate relation labels does not consistently improve model performance.** Specifically, macro $F_1$ increases when the number of candidate labels is expanded from top-3 to top-4, but slightly decreases at top-5. Interestingly, using the full relation label set achieves better performance than several top-k settings. A possible explanation is that the relation labels predicted by the SLM are not always accurate. When the ground-truth relation label is excluded from the top-k candidate set, the candidate space may incorrectly constrain the LLM. These results suggest that the effectiveness of candidate label selection depends not only on the number of selected labels but also on the prediction accuracy of the SLM.

**(3) Influence of the quality threshold on closed-source LLM-enhanced open-source LLMs**

Reducing the number of training samples is an effective strategy for lowering the computational and time costs associated with LLM adaptation. To investigate the impact of training data size on the KERE task, Fig.5c presents the macro $F_1$ scores obtained under different random sampling ratios based on one fold from the five-fold cross-validation setting.

Under random sample selection, the models achieve their lowest performance on the Abstract dataset when only 10% of the training data is retained. As the random sampling ratio increases from 10% to 20%, 50%, and 100%, macro $F_1$ generally shows an upward trend. This result suggests that randomly reducing the number of training samples may negatively affect the ability of models to learn relation patterns.

Based on these observations, this study further evaluates whether closed-source LLM-based sample quality assessment can improve sample selection. As shown in Fig.5d, when the quality threshold is set to 3, fewer than 50% of the training samples are retained. Nevertheless, the resulting macro $F_1$ reaches 66.87%, exceeding the 66.67% obtained by randomly selecting 50% of the training samples. **This result suggests that quality-based sample selection using a closed-source LLM can retain more informative training samples and partially mitigate the performance degradation caused by reducing the training data size.**

#### 4.2.3 Ethical statement on closed-source LLMs and computational cost analysis

Ethical considerations regarding the use of closed-source LLMs were taken into account in this study. All academic papers used in the experiments were obtained from publicly accessible open sources. Therefore, the experimental data did not involve the transmission of private or confidential scientific information to third-party LLM providers.

Table 11. GPU hours per epoch comparison across different models using NVIDIA 4090 24G GPU

| **Dataset** | **T5** | **LA** | **LA-C** | **BC** | **RE-SciBERT** |
|---|---|---|---|---|---|
| Abstract | 1 | 1 | 1.25 | 3 | **0.30** |
| Full-Text | 4 | 4 | 4.5 | 12 | **1** |

**Note.** Bold values indicate the lowest GPU hours per epoch.

To compare computational efficiency among representative models, this study selects T5, LLaMA2, LLAMA2-Chat, Baichuan2, and RE-SciBERT based on their performance across different evaluation metrics. T5 achieves strong macro recall and macro $F_1$, LLAMA2-Chat and Baichuan2 show advantages in macro precision, LLaMA2 performs well in micro $F_1$, and RE-SciBERT represents the best-performing SLM.

As shown in Table 11, the models exhibit clear differences in computational efficiency. RE-SciBERT requires the lowest computational cost, with 0.30 GPU hours on the Abstract dataset and 1 GPU hour on the Full-Text dataset. Among the LLMs, T5 and LLaMA2 require relatively low computational costs. In contrast, Baichuan2 requires approximately three times the GPU hours of T5 and LLaMA2 on both datasets.

**Overall, T5 provides a favorable balance between extraction performance and computational cost.** RE-SciBERT offers the highest computational efficiency but lower extraction performance than T5. Therefore, T5 appears to be a practical choice for KERE when both effectiveness and computational efficiency are considered.

### 4.3 Analysis of the ability of LLM-Based KERE models to mitigate SLM errors

To answer RQ2, this section examines whether LLM-based KERE models can reduce errors observed in SLM-based models such as RE-SciBERT. Beyond overall performance metrics, this study analyzes three error types manually identified described in Section 4.1.3, namely coreference resolution errors, relation direction prediction errors, and entity contextual semantic understanding errors. These error types are used to compare the error patterns of SLMs and LLM-based KERE models.

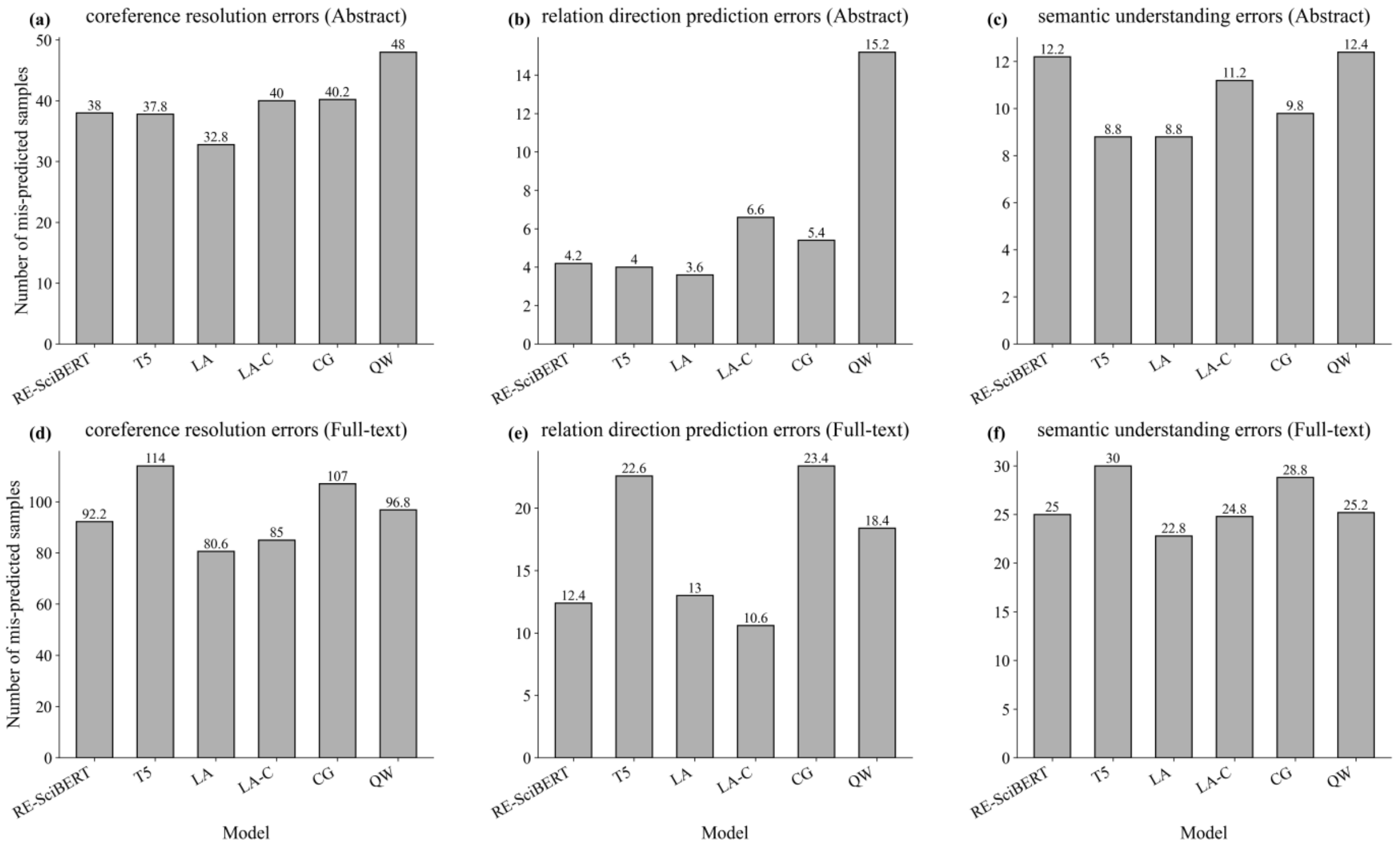

Fig 6. The distribution of errors in the prediction results of RE-SciBERT and different LLM-based models

### 4.3.1 Error types in the SLM-based KERE models

Coreference resolution errors occur when a model fails to correctly identify anaphoric relationships between entity mentions. Relation direction prediction errors occur when the direction of a directional relation is incorrectly identified. For instance, in a hypernym relationship between $X$ and $Y$, where $X$ is the hypernym and $Y$ is the hyponym, if the model incorrectly predicts $Y$ as the hypernym and $X$ as the hyponym, it results in a direction prediction error. Entity contextual semantic understanding errors occur when a model fails to capture the contextual relationship between two entities and consequently predicts an incorrect relation.

Different subsets of test samples are selected for the analysis of these errors. Cross-sentence samples are used to examine coreference resolution errors. Samples containing reversed entity pairs in directional relations, such as superclass-of and part-of, are used to analyze relation direction prediction errors. Samples in which the distance between two entities exceeds 250 characters are used to assess contextual semantic understanding, as identifying relations between distant entities requires greater use of contextual information.

### 4.3.2 Comparative analysis of error types across SLM and LLM-based KERE models

As shown in Fig.6, LLM-based KERE models generally make fewer errors than RE-SciBERT across the three error types. On the Abstract dataset, T5 and LLaMA2 show lower error rates in coreference resolution, relation direction prediction, and contextual semantic understanding. This suggests that LoRA-adapted open-source LLMs have advantages over the SLM in capturing coreferential relationships, directional information, and contextual semantics.

The results on the Full-Text dataset show a different pattern. LLaMA2 and LLaMA2-Chat achieve the lowest error rates for coreference resolution and contextual semantic understanding, while LLaMA2-Chat also makes the fewest relation direction prediction errors. In contrast, T5 produces more errors, particularly in coreference resolution and contextual semantic understanding.

This difference may be associated with the precision and recall patterns reported in Tables 5 and 6. T5 tends to achieve higher recall by predicting more potential relations, whereas LLaMA2 and LLaMA2-Chat achieve higher precision and make fewer predictions for uncertain relations. Full-text samples contain more complex contextual information than abstract samples. Thus, LLaMA2 and LLaMA2-Chat may help reduce incorrect relation predictions in complex contexts.

Table 12. Example in which T5 correctly identifies a directed comparison missed by RE-SciBERT

| Item | Content |
|---|---|
| Target entity pair | "G-Re-Rank" and "Struct-LP" |
| Input sentence | Using **Struct-LP**, the performance is significantly improved compared with the baseline, but not as well as **G-Re-Rank**. |
| Ground truth | directed-compare: G-Re-Rank performs better than Struct-LP |
| RE-SciBERT prediction | irrelevant |
| T5 prediction | directed-compare |
| T5-generated description | The relation label "directed-compare" indicates that the two phrases "G-Re-Rank" and "Struct-LP" are directly compared in terms of their performance. |

**Note.** Bold and underlined expressions in the examples indicate the entity mentions involved in the corresponding relation.

To provide a concrete illustration of how LLMs mitigate SLM errors, Table 12 presents a directed-compare example in which RE-SciBERT fails while the T5 model succeeds. For the target entity pair G-Re-Rank and Struct-LP, RE-SciBERT predicts irrelevant, whereas T5 predicts directed-compare, which is consistent with the ground-truth label. Although the first part of the sentence states that Struct-LP achieves an improvement over the baseline, the relation between the target entities is expressed in the subsequent contrastive clause, "but not as well as G-Re-Rank." This clause indicates that Struct-LP performs worse than G-Re-Rank and therefore establishes a directional comparison from G-Re-Rank to Struct-LP. The T5-generated description similarly identifies that the two methods are directly compared in terms of their performance.

This example is challenging because the sentence contains a third comparison target, namely the baseline. Moreover, the phrase "not as well as" expresses the performance comparison elliptically. RE-SciBERT fails to identify this implicit directional relation and classifies the entity pair as unrelated. In contrast, T5 correctly integrates the comparative and contrastive information to identify both the existence and direction of the relation. This case provides support for the error analysis in Fig.6, showing how T5 can mitigate relation direction prediction and entity contextual semantic understanding errors made by RE-SciBERT.

### 4.4 Analysis of optimization directions for LLM-based KERE models

To answer RQ3, this section analyzes prediction errors produced by LLM-based KERE models and the corresponding generated explanations to identify their limitations and explore potential directions for targeted optimization.

#### 4.4.1 Errors in LLM-based KERE models

This section analyzes prediction errors in LLM-based KERE models using the three error categories proposed by Das et al. (2022), namely spurious relations, missing relations, and wrong relations. A spurious relation occurs when an entity pair with the ground-truth label irrelevant is predicted as having a relation. A missing relation occurs when an entity pair with a valid relation is incorrectly predicted as irrelevant. A wrong relation refers to cases in which a non-irrelevant relation is assigned an incorrect relation type. Table 13 presents the distribution of these errors across different models.

Table 13. Errors in LLM-based KERE models

| model \ category | Abstract | | | Full-Text | | |
|---|---|---|---|---|---|---|
| | spurious | missing | wrong | spurious | missing | wrong |
| RE-SciBERT | 50 | 92.8 | 4.8 | 160.6 | 213.4 | 13 |
| R-BERT | 57.8 | 98 | 5.6 | 137.8 | 242.8 | 11 |
| T5 | 60.8 | **55** | 4.4 | 281.8 | **113.6** | 8.6 |
| LA | **41.8** | 79.2 | **3** | 119.8 | 228.2 | **5.4** |
| LA-C | 49.8 | 110.8 | 3.2 | **117.8** | 229.8 | 6 |
| CG | 44.4 | 130.6 | 4.4 | 236.2 | 219.6 | 26.6 |
| QW | 74 | 138.8 | 4 | 169.4 | 264.2 | 6 |

**Note.** Bold values indicate the lowest number of errors among the evaluated models for the corresponding error type and dataset.

As shown in Table 13, wrong relations are the least frequent error type across all models, indicating that models generally avoid incorrectly assigning wrong relation labels to entity pairs. For decoder-only LLMs such as LLaMA2 and GLM2, missing relations are more frequent than spurious relations, suggesting that these models are more likely to classify valid relations as irrelevant. In contrast, for the encoder-decoder model T5, spurious relations are more frequent than missing relations, indicating a greater tendency to assign relation labels to irrelevant entity pairs.

Table 14. Explanations generated by the fine-tuned LLMs

| id | category | sample | result | explanation |
|---|---|---|---|---|
| 1 | spurious | The resulting logical expression is then transformed by a **<u>planning algorithm</u>** into efficient **<u>query optimization</u>** in a relational database. | **Gold**: irrelevant **Prediction**: used-for | The planning algorithm is used to transform the original query into an efficient program which then executes the program to obtain the desired result. |
| 2 | missing | In this paper, we present an **<u>unlexicalized parser for German</u>** which employs smoothing and suffix analysis to achieve a labelled bracket F-score of 76.2, higher than previously reported results on the **<u>NEGRA corpus</u>**. | **Gold**: with-dataset **Prediction**: irrelevant | The two phrases 'unlexicalized parser for Germa' and 'NEGRA corpus' do not have any relation as both are independent of each other. |
| 3 | wrong | This **<u>statistical approach</u>** aims to **<u>minimize expected loss of translation errors under loss functions</u>** that measure translation performance. | **Gold**: used-for **Prediction**: part-of | The phrase 'statistical approach' is part-of the term 'minimize expected loss of translation errors under loss functions' because it is used to perform the minimization. |

**Note.** Bold and underlined expressions in the examples indicate the entity mentions involved in the corresponding relation.

To further examine the prediction behavior of LLMs, this study analyzes the explanations generated for incorrectly predicted samples. Representative examples are presented in Table 14. In Sample 1, the model captures the general meaning of the text but fails to focus on the target entity pair, resulting in an incorrect relation prediction. In Sample 2, the model predicts that no relation exists between the two entities. The large textual distance between them make their semantic connection more difficult to identify. In Sample 3, the model incorrectly captures the semantics of the sample and predicts a part-of relation instead of the correct used-for relation.

#### 4.4.2 Design of optimization directions

Table 15. Results of ensemble method

| **Metrics** (%) / **Model** | | **Abstract** | | | **Full-Text** | | |
|---|---|---|---|---|---|---|---|
| | | LA | T5 | LA +T5 | LA | T5 | LA +T5 |
| $F_1$ | used-for | 84.79 | 85.31 | **85.74** | 87.27 | **88.22** | 87.82 |
| | superclass-of | 70.67 | 73.80 | **75.03** | 74.22 | 73.32 | **76.05** |
| | part-of | 71.20 | 76.21 | **76.22** | 73.48 | **73.89** | 73.72 |
| | sequence-of | 65.79 | 67.26 | **67.91** | 75.43 | 77.09 | **77.34** |
| | with-metric | 65.44 | 71.55 | **73.08** | 58.30 | 61.41 | **62.25** |
| | compare | 74.53 | **76.59** | **76.59** | **78.76** | 77.23 | 76.58 |
| | directed-compare | 53.23 | **60.35** | 58.97 | 51.85 | 56.93 | **57.93** |
| | with-dataset | 46.57 | 46.74 | **59.13** | 10.00 | **34.02** | **34.02** |
| | problem-exist | 25.00 | **87.50** | **87.50** | 51.64 | 61.51 | **61.91** |
| | irrelevant | 97.86 | **97.94** | 97.88 | **97.97** | 97.66 | 97.63 |
| Macro P | | 77.38±8.03 | **75.24**±5.57 | 73.90±4.98 | **74.12**±2.09 | 64.72±2.63 | 64.46±2.30 |
| Macro R | | 60.94±7.22 | 75.08±4.94 | **77.04**±5.13 | 60.95±2.02 | 79.45±4.29 | **80.56**±4.32 |
| Macro $F_1$ | | 66.47±6.14 | 74.08±3.30 | **74.32**±3.40 | 65.89±1.50 | 70.13±2.32 | **70.53**±2.05 |
| Macro $F_1$ (w/o irrelevant) | | 62.91±6.74 | 71.37±3.61 | **73.37**±0.98 | 62.33±1.66 | 67.07±2.57 | **67.51**±2.28 |
| Micro $F_1$ | | 96.02±0.51 | **96.15**±0.50 | 96.06±0.50 | **96.27**±0.11 | 95.74±0.13 | 95.69±0.12 |

**Note.** Bold values indicate the best performance among the models compared for the corresponding metric on the Abstract dataset and the Full-Text dataset.

As shown in Table 13, decoder-only models such as LLaMA2 are less prone to spurious relation errors, whereas encoder-decoder models such as T5 are less likely to produce missing relation errors. To combine the strengths of these models, this study proposes a rule-based ensemble strategy based on their complementary error patterns. The ensemble uses T5 as the base model and incorporates the predictions of LLaMA2 to revise its outputs. For each sample, predictions are first obtained from both T5 and LLaMA2. If T5 predicts irrelevant while LLaMA2 predicts a non-irrelevant relation, the LLaMA2 prediction is adopted. When both models predict non-irrelevant relations but assign different relation types, the LLaMA2 prediction is also used to replace the T5 output. In all other cases, the T5 prediction is retained.

As shown in Table 15, the ensemble method achieves a higher macro $F_1$ than individual models. The ensemble improves macro recall but results in lower macro precision. This pattern indicates that the combined model identifies more valid relations, thereby reducing missing relation errors, while introducing additional false positive predictions.

What's more, based on the analysis of explanations generated by the LoRA-adapted open-source LLMs, several optimization directions can be identified. First, when the model fails to focus on the target entity pair, future research can optimize the prompt information. For example, special symbols can be used to mark the entities, or the specific positions of the entities in the samples can be clearly indicated. Second, for entity pairs separated by long textual distances, LLMs can be used to rewrite samples to change the distance between entities. Finally, when the model incorrectly captures the contextual semantics of a sample, the retrieval-augmented generation method can be adopted. In this way, relevant content related to the samples can be extracted from academic papers, and these contents can be used to expand the original samples.

# 5. Conclusion and future works

This study investigates five LLM-based strategies for KERE on academic papers and addresses three research questions. For RQ1, the results in Section 4.2 show differences across model types and datasets. LoRA-adapted open-source LLMs achieve the best overall performance, with T5 obtaining the highest macro F1 on both the Abstract and Full-Text datasets. Closed-source LLM-based CoT prompting performs less effectively, although GPT-5 mini outperforms Qwen-turbo. LLM-enhanced SLMs benefit from semantic augmentation, and SLM-enhanced closed-source LLMs gain from candidate relation selection. By contrast, SLM-enhanced open-source LLMs do not outperform their corresponding LoRA-adapted models. For closed-source LLM-enhanced open-source LLMs, relation explanation augmentation has different effects across models and datasets, improving Qwen2.5 on the Abstract dataset but not T5. LLM-based sample quality selection also does not directly improve overall performance, although the ablation results show that it can outperform random sampling. In addition, performance differences between the Abstract and Full-Text datasets vary across models. Although T5 and some other models achieve lower macro $F_1$ scores on the Full-Text dataset, several SLMs perform better on Full-Text dataset. These differences may result from interactions among model architecture, sample composition, and the distribution of relation types.

For RQ2, the analysis in Section 4.3 shows that LLM-based KERE models can reduce several error types found in SLM-based models, including coreference resolution errors, relation direction prediction errors, and entity contextual semantic understanding errors. The extent of improvement varies across models and datasets. LLaMA2 and LLaMA2-Chat perform well in full-text contexts, whereas T5 tends to achieve higher recall but also introduces more incorrect relation predictions.

For RQ3, the analysis in Section 4.4 identifies three main limitations of LLM-based KERE models, namely insufficient focus on target entity pairs, difficulty in identifying relations between distant entities, and failure to correctly capture the contextual semantics of a sample. Based on these observations, several optimization directions are proposed, including explicit entity marking, context rewriting, retrieval-augmented generation, and ensemble strategies that combine models with different error patterns.

These findings have several theoretical implications. First, the results suggest that task-specific alignment may be more important than relying solely on the general reasoning capabilities of a model for KERE. Although closed-source LLMs possess broad linguistic knowledge and can generate reasoning processes through CoT prompting, they perform less effectively than the LoRA-adapted T5 model. This finding indicates that fine-grained relation extraction depends substantially on learning from the target relation schema and manually annotated data. Second, the experimental results on the Abstract and Full-Text datasets show that some models perform better on the Abstract dataset, whereas others achieve comparable or higher performance on the Full-Text dataset. This finding indicates that KERE performance is shaped by interactions among model architecture, sample composition, and relation-type distribution, rather than by a fixed difficulty hierarchy between samples derived from abstracts and those derived from full-text articles. Therefore, cross-dataset analyses should focus on model robustness and performance variations across relation types. Finally, the different precision value, recall value and error patterns indicate that model architectures and enhancement strategies produce different prediction behaviors. T5 tends to identify more potential relations, resulting in higher recall but also more spurious relations. In contrast, decoder-only models are generally more conservative and may therefore miss valid relations. In addition, LLM-generated entity descriptions can improve SLM performance. These findings suggest that collaboration between models is more likely to be effective when the additional information compensates for the limitations of the receiving model.

These findings also have several practical implications. First, model selection should be aligned with the requirements of the intended application. LoRA-adapted T5 offers a favorable trade-off between extraction performance and computational cost, making it suitable when comprehensive coverage across relation types is required. RE-SciBERT remains a practical option in resource-constrained settings, whereas models that achieve higher precision, such as LLaMA2 and LLaMA2-Chat, may be preferable when false positives are particularly costly. Conversely, T5's high recall makes it well suited to candidate relation discovery, where the extracted relations can subsequently be reviewed by humans. Second, before deployment, models should be validated using samples that reflect the intended application scenario. Because models do not perform consistently across the Abstract and Full-Text datasets, an advantage observed on abstract samples cannot be directly generalized to samples derived from full-text articles, and vice versa. When an application involves both abstracts and full-text data, model robustness across different sources of academic text should be carefully evaluated. Third, model-enhancement strategies should be applied selectively, because additional information does not necessarily improve performance. For example, SLM-predicted candidate labels should be used as soft guidance rather than as hard constraints. Otherwise, an inaccurate candidate set may exclude the ground-truth relation and prevent the LLM from making the correct prediction.

In future work, this study plans to develop LLM-integrated KERE methods based on the optimization directions identified in this study. The effectiveness of these methods will be further evaluated on datasets from a wider range of academic domains to examine their generalizability. In addition, this study used LLM-generated explanations to provide additional insights into the causes of incorrect predictions. Future research will combine machine learning interpretability techniques with LLMs to further investigate the textual information that models rely on during KERE and provide a deeper understanding of their prediction behavior.

## Declarations

The authors declared no potential conflicts of interest with respect to the research, authorship, and publication of this article.

## Acknowledgments

This work is supported by National Natural Science Foundation of China (Grant No.72074113), Jiangsu Social Science Fund (Grant No. 26XZA001), and the 2025 Higher Education Teaching Reform Research Project of Soochow University.